\documentclass[runningheads]{llncs}
\usepackage[T1]{fontenc}
\usepackage{graphicx,verbatim}
\usepackage{hyperref}
\usepackage{amsmath}
\usepackage{amssymb}
\usepackage{booktabs}
\usepackage{tabularx}
\newcolumntype{C}{>{\centering\arraybackslash}X}
\usepackage{subcaption}
\usepackage{booktabs}
\usepackage{array}
\usepackage{graphicx}
\usepackage{subcaption}
\usepackage{xcolor}
\usepackage{enumitem}
\usepackage{caption}

\usepackage{marvosym}

\newcommand{\modelname}{MIRAGE}

\begin{document}

\title{MIRAGE: Knowledge Graph-Guided Cross-Cohort MRI Synthesis \\for Alzheimer’s Disease Prediction}

\author{Guanchen Wu\inst{1} \and
Zhe Huang\inst{2} \and
Yuzhang Xie\inst{1} \and
Runze Yan\inst{1} \and
Akul Chopra\inst{1} \and
Deqiang Qiu\inst{1} \and
Xiao Hu\inst{1} \and
Fei Wang\inst{2} \and
% Carl Yang\inst{1}\thanks{Corresponding author: j.carlyang@emory.edu}
Carl Yang\inst{1}\Letter
}

\authorrunning{G. Wu et al.}

\institute{Emory University, Atlanta, USA \and
Cornell University, New York, USA \\
\email{j.carlyang@emory.edu} % 通常通讯邮箱放在这里
}

% \institute{Emory University \and
% Cornell University
% % \email{j.carlyang@emory.edu}\\
% % \url{http://www.springer.com/gp/computer-science/lncs} \and
% % ABC Institute, Rupert-Karls-University Heidelberg, Heidelberg, Germany\\
% % \email{\{abc,lncs\}@uni-heidelberg.de}
% }

% \author{Anonymized Authors}  %% Added for anonymized MICCAI submission
% \authorrunning{Anonymized Author et al.}
% \institute{Anonymized Affiliations \\
%     \email{email@anonymized.com}}
  
\maketitle              % typeset the header of the contribution
\begin{abstract}

% Magnetic resonance imaging (MRI) provides critical structural information for Alzheimer’s disease (AD) diagnosis. However, MRI acquisition is costly, and many real-world clinical cohorts contain only electronic health records (EHR). Although EHR data are widely available, building a robust EHR-based AD prediction model remains challenging due to their noisy, high-dimensional, and semi-structured nature. In this work, we propose \modelname, an image generation framework that synthesizes MRI scans for EHR-only cohorts by leveraging a biomedical knowledge graph (KG). The KG connects heterogeneous clinical variables by mapping diverse EHR features to shared biomedical concepts, enabling cross-patient information transfer and guiding the propagation of disease-related biomarkers from MRI-available cohorts to EHR-only cohorts. \modelname\ integrates KG-guided embedding propagation with distribution alignment to ensure that synthesized MRI representations are both biologically meaningful and consistent with real MRI distributions. Experimental results demonstrate that the synthesized MRIs preserve AD-related biomarkers and improve downstream AD classification in EHR-only cohorts.
Reliable Alzheimer's disease (AD) diagnosis increasingly relies on multimodal assessments combining structural Magnetic Resonance Imaging (MRI) and Electronic Health Records (EHR). However, deploying these models is bottlenecked by modality missingness, as MRI scans are expensive and frequently unavailable in many patient cohorts. Furthermore, synthesizing de novo 3D anatomical scans from sparse, high-dimensional tabular records is technically challenging and poses severe clinical risks. To address this, we introduce MIRAGE, a novel framework that reframes the missing-MRI problem as an anatomy-guided cross-modal latent distillation task. First, MIRAGE leverages a Biomedical Knowledge Graph (KG) and Graph Attention Networks to map heterogeneous EHR variables into a unified embedding space that can be propagated from cohorts with real MRIs to cohorts without them. To bridge the semantic gap and enforce physical spatial awareness, we employ a frozen pre-trained 3D U-Net decoder strictly as an auxiliary regularization engine. Supported by a novel cohort-aggregated skip feature compensation strategy, this decoder acts as a rigorous structural penalty, forcing 1D latent representations to encode biologically plausible, macro-level pathological semantics. By exclusively utilizing this distilled ``diagnostic-surrogate'' representation during inference, MIRAGE completely bypasses computationally expensive 3D voxel reconstruction. Experiments demonstrate that our framework successfully bridges the missing-modality gap, improving the AD classification rate by 13\% compared to unimodal baselines in cohorts without real MRIs.
\footnote{The implementation details and codes are available in the anonymous repository at \href{https://anonymous.4open.science/r/MIRAGE-71B6}{https://anonymous.4open.science/r/MIRAGE-71B6}.}

% Authors must provide keywords and are not allowed to remove this Keyword section.

\end{abstract}

\section{Introduction}
\label{sec:intro}

Alzheimer's disease (AD) is a progressive neurodegenerative disorder requiring early diagnosis for effective intervention \cite{barker2002relative}. In clinical practice, reliable AD diagnosis relies on comprehensive assessment combining structural neuroimaging, such as Magnetic Resonance Imaging (MRI), and Electronic Health Records (EHR)\cite{frisoni2025new,fu2025identifying}. These data sources provide complementary value: MRI identifies structural biomarkers such as hippocampal atrophy, while EHR captures clinical trajectories including cognitive assessments, medical history, and demographics \cite{qiu2022multimodal,xie2022survival,zhang2024tacco}. Recognizing this synergy, state-of-the-art machine learning models for AD classification increasingly employ multimodal architectures \cite{qiu2022multimodal,liu2025multi}. By jointly leveraging EHR and MRI,  models achieve higher reliability and accuracy than unimodal approaches, establishing a new standard for AD analysis.

However, fundamental modality imbalance limits real-world multimodal model deployment \cite{zhang2022m3care,le2025multimodal}. Although EHR is routinely collected and accessible, expensive and clinically complex MRI acquisition \cite{lin2024designing,edelstein2010mri} causes frequent missing scans in large cohorts. Because patients with and without MRI frequently originate from partially overlapping or institution-specific cohorts under varying protocols, this missing-modality challenge becomes a cross-cohort knowledge transfer problem where MRI-rich cohorts structurally guide EHR-only ones. Here, EHR-only patients combine clinical records with transferred structural guidance to generate MRI representations.
Synthesizing patient-specific 3D MRIs from tabular records is a natural but ill-posed solution: generating anatomically faithful voxel-level scans from low-dimensional EHR remains challenging and clinically risky \cite{duenias2025hyperfusion,zhu2023make}. Crucially, downstream diagnostic models require MRI-encoded structural priors and latent disease manifolds rather than visually perfect images. Therefore, we reformulate the problem as anatomy-guided cross-modal distillation. Instead of reconstructing full MRI volumes at inference, we project EHR into a structurally grounded latent space to activate MRI-derived priors directly in feature space, avoiding expensive, artifact-prone voxel reconstruction.

Extracting a diagnostic-quality structural surrogate from EHR without empirical neuroimaging faces three challenges: (i) \textit{EHR Sparsity and Heterogeneity:} Raw electronic health records are high-dimensional, sparse, and affected by schema variation across institutions. Direct reliance on this data obscures neurodegenerative states and confounds representation learning; (ii) \textit{The Cross-Modal Semantic Gap:} Direct mapping from sparse, unstructured clinical features to a complex 3D imaging distribution often leads to posterior collapse and fails to capture structural priors; (iii) \textit{Lack of Anatomical Constraints:} Standard cross-modal mappings lack spatial awareness. Projecting clinical data into an imaging manifold without explicit structural bounds yields biologically implausible latent representations, reducing prognostic value for AD.

% To address these challenges, we introduce MIRAGE, a cross-cohort inference framework designed to transfer MRI-derived structural knowledge from imaging-rich cohorts to EHR-only cohorts. MIRAGE decouples learning into graph-guided semantic propagation and anatomy-constrained joint optimization. First, to overcome EHR sparsity, Biomedical Knowledge Graphs (KGs) \cite{huang2020fusion,jiang2023graphcare,wang2022knowledge,xie2025hypkg} and a Graph Attention Network (GAT) \cite{velivckovic2017graph,brody2021attentive} map diverse clinical variables to a unified semantic space, inferring missing contexts from data-rich nodes. To bridge the semantic gap and enforce spatial awareness, a pre-trained 3D U-Net decoder is used as an auxiliary regularization engine. During training, an adapter maps EHR embeddings into MRI latent space. To satisfy the frozen 3D decoder requirement for high-frequency details, we introduce a Cohort-Aggregated Skip Feature Compensation strategy. By retrieving spatial priors (skip connections) from the top-$K$ clinically analogous patients, the decoder reconstructs a pseudo-3D MRI to compute voxel-level penalties (e.g., SSIM \cite{wang2004image}). This reconstruction is not intended to synthesize exact patient anatomy; instead, it acts as a structural penalty that forces 1D latent representations to encode macro-level pathological semantics. During inference, this heavy 3D generative branch is discarded. A classification head processes the distilled 1D latent vector, enabling efficient and accurate AD screening for EHR-only patients.

To address these challenges, we introduce MIRAGE, a cross-cohort inference framework transferring MRI-derived structural knowledge from imaging-rich to EHR-only cohorts. MIRAGE decouples learning into graph-guided semantic propagation and anatomy-constrained joint optimization. To overcome EHR sparsity, Biomedical Knowledge Graphs (KGs) \cite{huang2020fusion,jiang2023graphcare,wang2022knowledge,xie2025hypkg} and a Graph Attention Network (GAT) \cite{velivckovic2017graph,brody2021attentive} map diverse clinical variables to a unified semantic space, inferring missing contexts from data-rich nodes. To bridge semantic gaps and enforce spatial awareness, a pre-trained 3D U-Net decoder serves as an auxiliary regularization engine. During training, an adapter maps EHR embeddings into MRI latent space. To satisfy the frozen 3D decoder's requirement for high-frequency details, we introduce Cohort-Aggregated Skip Feature Compensation. By retrieving spatial priors (skip connections) from top-$K$ clinically analogous patients, the decoder reconstructs pseudo-3D MRIs to compute voxel-level penalties (e.g., SSIM \cite{wang2004image}). Rather than synthesizing exact patient anatomy, this reconstruction acts as a structural penalty forcing 1D latent representations to encode macro-level pathological semantics. Discarding this heavy 3D generative branch during inference, a classification head processes the distilled 1D latent vector, enabling efficient and accurate AD screening for EHR-only patients.

To the best of our knowledge, MIRAGE is the first framework to cast the missing-MRI problem as an anatomy-guided latent distillation task using a frozen 3D autoencoder as a structural regularizer. Our contributions are summarized as follows: (i) An Anatomy-Guided Cross-Modal Distillation framework that refines EHR representations through latent regularization, showing that structural priors improve AD diagnosis without pixel-level synthesis. (ii) A KG-Driven Cohort Alignment mechanism that addresses inconsistent EHR schemas by linking heterogeneous cohorts into a unified, semantically dense graph. (iii) An Auxiliary Decoding Regularization strategy that leverages cohort-retrieved skip features to bridge macro-prognostic representations and micro-texture constraints of a 3D MRI decoder. (iv) Clinical validation demonstrating that the framework bridges the missing-modality gap, improving the AD classification rate by 13\% over unimodal baselines while maintaining high inference efficiency.

\section{Method}

% \paragraph{\textbf{Problem Formulation.}} Let $KG = (X, R, RT)$ be a heterogeneous biomedical knowledge graph, where nodes $X = \mathcal{P}_{\mathrm{tr}} \cup \mathcal{P}_{\mathrm{te}} \cup \mathcal{M}$ represents training patients, testing patients, and medical concepts, respectively. For each $i \in \mathcal{P}_{\mathrm{tr}}$, we observe a 3D MRI scan $X_i \in \mathbb{R}^{H \times W \times D}$ and a diagnostic label $y_i$. For $j \in \mathcal{P}_{\mathrm{te}}$, only EHR features are available. Our goal is to synthesize an MRI volume $\widehat{X}_j$ that preserves realistic anatomical structures and maintains diagnostic utility for downstream classification.

\begin{figure*}[htbp!]
\centering
\captionsetup{font=scriptsize} 
\includegraphics[width=0.75\textwidth]{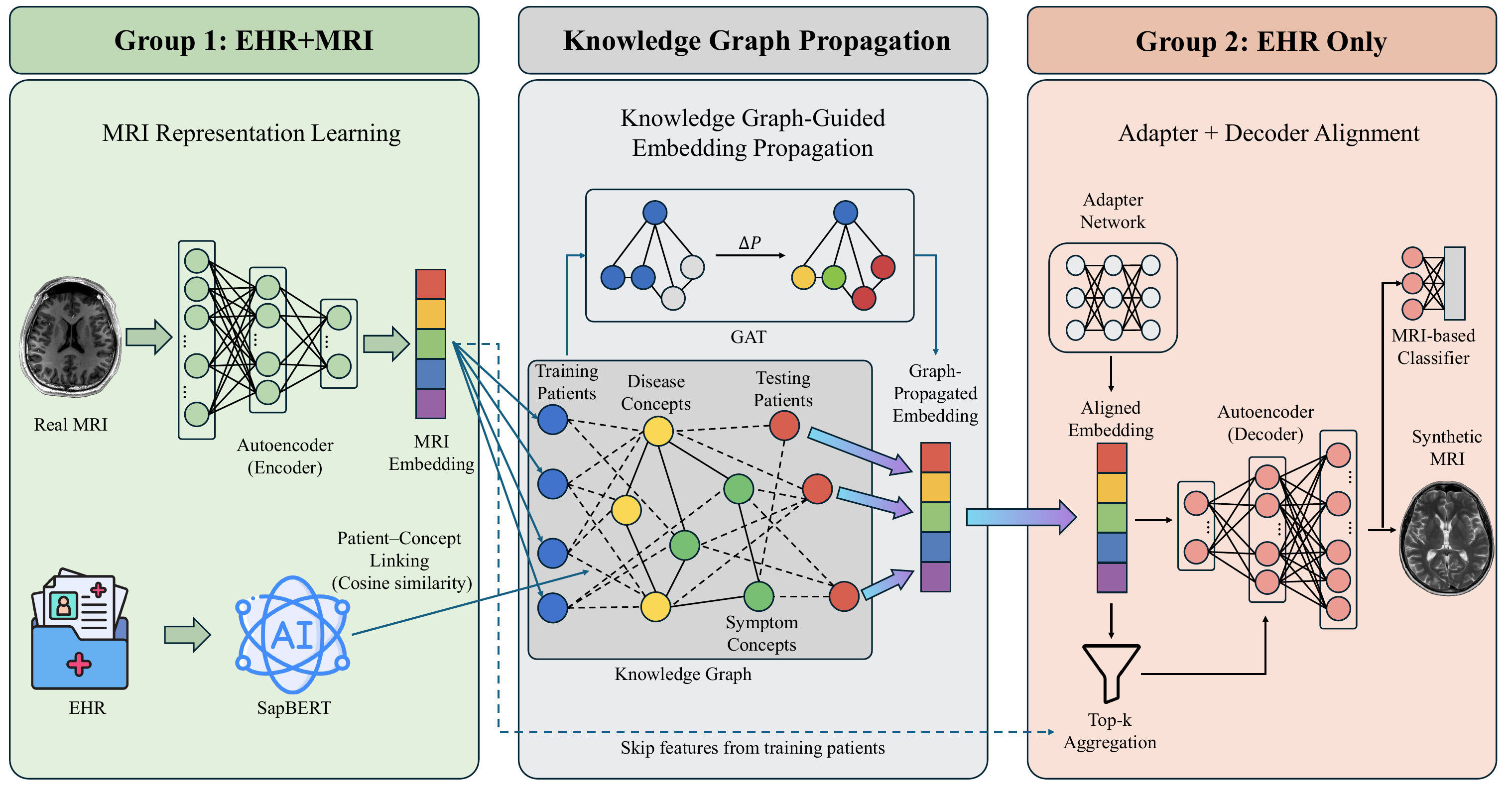}
\caption{Overall framework of \modelname. 
% Real MRIs are encoded into latent embeddings and skip features by a 3D U-Net autoencoder. A knowledge graph with GAT propagates MRI information to EHR-only patients to produce synthetic embeddings, which are aligned by an adapter and decoded with top-$k$ skip features to generate synthetic MRI.
}
\label{fig:framework}
\end{figure*}

\paragraph{\textbf{Problem Formulation and Framework Overview.}} Let $KG = (X, R, RT)$ be a heterogeneous biomedical knowledge graph, where nodes $X = \mathcal{P}{\mathrm{tr}} \cup \mathcal{P}{\mathrm{te}} \cup \mathcal{M}$ denote training patients, testing patients, and medical concepts. For each $i \in \mathcal{P}{\mathrm{tr}}$, we observe a 3D MRI scan $X_i \in \mathbb{R}^{H \times W \times D}$ and label $y_i$. For $j \in \mathcal{P}{\mathrm{te}}$, only EHR features are available. The objective is to synthesize an MRI volume $\widehat{X}_j$ preserving anatomical realism and diagnostic utility for downstream classification.
To bridge the modality gap for patients without neuroimaging, we propose \modelname\ (\underline{M}RI \underline{I}nference and \underline{R}econstruction via \underline{A}ligned \underline{G}raph-guided \underline{E}mbeddings). The framework integrates representation learning, graph-based propagation, and generative decoding. As shown in Figure~\ref{fig:framework}, it operates in three phases. First, a 3D autoencoder maps real MRIs to a compact latent space. Second, a GAT propagates these representations over the biomedical knowledge graph to estimate latent embeddings for EHR-only patients. Finally, Adapter+Decoder alignment refines these embeddings and reconstructs synthetic MRI volumes while optimizing diagnostic utility.

\paragraph{\textbf{MRI Representation Learning.}} The framework learns a compact latent representation of neurological structures using a 3D U-Net autoencoder (AE) with encoder $f_{\mathrm{enc}}$ and decoder $f_{\mathrm{dec}}$, trained on patients with real neuroimaging data: ${X_i \in \mathbb{R}^{1 \times H \times W \times D} : i \in \mathcal{P}{\mathrm{tr}}}$.
The encoder $f{\mathrm{enc}}$ compresses input $X_i$ into a latent manifold $z_i \in \mathbb{R}^d$ and extracts hierarchical spatial features $\text{skip}i = {\text{skip}i^{(1)}, \dots, \text{skip}i^{(4)}}$. The decoder $f{\mathrm{dec}}$ uses the latent vector and skip connections to reconstruct the volume:
\begin{equation}
z_i = f{\mathrm{enc}}(X_i), \quad \widehat{X}i = f{\mathrm{dec}}(z_i, \text{skip}i).
\end{equation}
Skip connections preserve high-frequency anatomical details. The autoencoder minimizes voxel-level mean squared error (MSE):
\begin{equation}
\mathcal{L}{\text{AE}} = \frac{1}{|\mathcal{P}{\mathrm{tr}}|} \sum_{i \in \mathcal{P}{\mathrm{tr}}} | f{\mathrm{dec}}(f_{\mathrm{enc}}(X_i), \text{skip}_i) - X_i |2^2.
\end{equation}
After convergence, decoder weights $f{\mathrm{dec}}$ are frozen and used as the generative engine for MRI synthesis in later stages.

\paragraph{\textbf{KG-Guided Embedding Propagation.}}
% With the MRI latent space established, the next challenge is to infer corresponding latent representations for testing patients who only possess EHR data. We formulate this as an embedding propagation problem over a heterogeneous biomedical knowledge graph with following steps: 
With the MRI latent space established, the next step is to infer latent representations for testing patients with only EHR data. We formulate this as embedding propagation over a heterogeneous biomedical knowledge graph.

\textbf{(I) Patient-to-KG Linking via EHR Concepts.}
% We leverage a medical knowledge graph anchored by concept nodes $\mathcal{M}$ representing established clinical ontology. Every patient $p \in \mathcal{P}_{\mathrm{tr}} \cup \mathcal{P}_{\mathrm{te}}$ is injected into this graph as a distinct node. To establish edges between patient nodes and medical concepts, we utilize SapBERT \cite{liu2021self}, a state-of-the-art biomedical language model \cite{xie2024promptlink}. SapBERT generates 768-D semantic embeddings for both the textual descriptions of EHR concepts and the KG nodes. 
% By computing the cosine similarity between these embeddings, we draw edges directly connecting the new patient nodes to the existing KG concept nodes, effectively grounding the clinical profiles within the broader biomedical ontology.
We use a medical knowledge graph with concept nodes $\mathcal{M}$ representing clinical ontology. Each patient $p \in \mathcal{P}{\mathrm{tr}} \cup \mathcal{P}{\mathrm{te}}$ is added as a node. Edges between patients and concepts are created using SapBERT \cite{liu2021self,xie2024promptlink,wu2025utilizing}, which generates 768-D embeddings for EHR concept descriptions and KG nodes. Cosine similarity connects patient nodes to concept nodes, grounding clinical profiles in the biomedical ontology. 

\textbf{(II) Two-Stage GAT Training.}
% Node features $x^{(0)}$ are initialized as $z_i$ (AE latent vector) for $x \in \mathcal{P}_{\mathrm{tr}}$, $g_x$ (SapBERT embedding) for $x \in \mathcal{M}$, and $\mathbf{0}$ for testing patients $x \in \mathcal{P}_{\mathrm{te}}$. Because these initial features reside in heterogeneous latent spaces with varying dimensions, we apply a node-type-specific linear transformation to project them into a shared $d$-dimensional embedding space before message passing. Specifically, the aligned initial state $h_x^{(0)}$ for node $x$ is obtained via $h_x^{(0)} = W_{\tau(x)} x^{(0)} + b_{\tau(x)}$, where $W_{\tau(x)}$ and $b_{\tau(x)}$ are the learnable weight matrix and bias for node type $\tau(x) \in \{\text{Patient}, \text{Concept}\}$. Consequently, testing patients initially hold no distinct internal features, meaning their representations are synthesized entirely through the aggregation of topological information from their linked medical concept nodes during the GAT propagation.
% We apply a three-layer \texttt{GATv2} \cite{brody2021attentive} with residual connections to allow dynamic, attention-weighted feature aggregation across the graph: $x^{(\ell + 1)} = \mathrm{GATv2}\!\left(x^{(\ell)}, \mathcal{E}\right)$. Training the GAT is conducted in two stages to ensure both structural integrity and generative compatibility.
Node features $x^{(0)}$ are initialized as $z_i$ for $x \in \mathcal{P}{\mathrm{tr}}$, $g_x$ for $x \in \mathcal{M}$, and $\mathbf{0}$ for $x \in \mathcal{P}{\mathrm{te}}$. Because features lie in heterogeneous spaces, node-type-specific linear layers project them into a shared $d$-dimensional space: $h_x^{(0)} = W_{\tau(x)} x^{(0)} + b_{\tau(x)}$. Testing patients initially contain no internal features and rely on aggregation from linked concept nodes. We apply a three-layer GATv2 \cite{brody2021attentive} with residual connections for attention-weighted aggregation: $x^{(\ell + 1)} = \mathrm{GATv2}\left(x^{(\ell)}, \mathcal{E}\right)$. Training proceeds in two stages.

\indent \textit{Stage 1: Supervised Regression.} 
The network first regresses ground-truth MRI latent vectors ($z_i$):
% The network is first trained to regress the ground-truth structural MRI latent vectors ($z_i$) extracted by the pre-trained Autoencoder for the training cohort:
\begin{equation}
\mathcal{L}_{\mathrm{GCN}} = \sum_{i \in \mathcal{P}_{\mathrm{tr}}} \| h_i^{(L)} - z_i \|^2.
\end{equation}

\textit{Stage 2: Decoder-Consistency Fine-Tuning.} 
To ensure compatibility with the frozen AE decoder, GAT outputs $h_i^{(L)}$ are decoded and penalized for reconstruction errors using SSIM:
% To guarantee that the graph-derived embeddings are fully compatible with the frozen AE decoder, we introduce a secondary fine-tuning phase. The GAT outputs $h_i^{(L)}$ are passed through $f_{\mathrm{dec}}$, and the network is penalized for visual reconstruction errors. We incorporate the Structural Similarity Index Measure (SSIM) to capture perceptual differences in local luminance, contrast, and structure:
\begin{equation}
\mathcal{L}_{\text{recon\_GAT}} = \| f_{\mathrm{dec}}(h_i^{(L)}) - X_i \|^2 + (1 - \mathrm{SSIM}(f_{\mathrm{dec}}(h_i^{(L)}), X_i)).
\end{equation}
% Following this two-stage optimization, the GAT successfully infers synthetic embeddings $u_j = h_j^{(L)}$ for all testing patients $j \in \mathcal{P}_{\mathrm{te}}$.
After this optimization, the GAT infers synthetic embeddings $u_j = h_j^{(L)}$ for all testing patients $j \in \mathcal{P}_{\mathrm{te}}$.

\paragraph{\textbf{Adapter and Decoder Alignment.}}
% While the GAT provides strong approximations of the latent space, slight distributional shifts can degrade the final MRI synthesis. To correct this, we introduce a trainable adapter network and a diagnostic classification head, which are jointly optimized alongside the frozen decoder. 
Although GAT provides strong latent approximations, distributional shifts degrade MRI synthesis. We introduce a trainable adapter and classification head optimized with the frozen decoder.

\textbf{(I) WC Transformer.} 
% Prior to adaptation, we align the distribution of the graph-propagated embeddings ($u_i$) with that of the true AE latent space using a Whitening-Coloring (WC) transformation \cite{li2017universal}. By calculating the means ($\mu_u, \mu_z$) and covariance matrices ($\Sigma_u, \Sigma_z$) of the propagated and target latent distributions respectively, we obtain the statistically aligned embedding 
Before adaptation, we align graph-propagated embeddings ($u_i$) with the AE latent space using a Whitening-Coloring (WC) transformation \cite{li2017universal}. Using means ($\mu_u, \mu_z$) and covariance matrices ($\Sigma_u, \Sigma_z$), we obtain the aligned embedding $\tilde{u}_i$:
$$\tilde{u}_i = \Sigma_z^{1/2} \Sigma_u^{-1/2} (u_i - \mu_u) + \mu_z$$
% This explicit alignment of covariance structures is crucial for improving decoding stability and overall MRI synthesis quality when generating features for the fixed AE decoder.
This alignment improves decoding stability and synthesis quality for the decoder.

\textbf{(II) Adapter Network and Skip Feature Aggregation.}
% The adapter $A_\theta$ maps the GAT-inferred embedding $u_j$ strictly into the decoder-compatible space: $\tilde{z}_j = A_\theta(u_j)$. However, decoding high-resolution MRIs also requires spatial skip features, which are naturally missing for testing patients. We resolve this via a top-$K$ nearest-neighbor aggregation strategy. For a testing embedding $u_j$, we identify the $K$ most similar training patients based on cosine similarity in the latent space. Their corresponding skip features are aggregated using normalized similarity weights:
The adapter $A_\theta$ maps GAT embeddings $u_j$ into the decoder space: $\tilde{z}j = A\theta(u_j)$. High-resolution decoding also requires spatial skip features missing for testing patients. We address this with top-$K$ nearest-neighbor aggregation. For embedding $u_j$, the $K$ most similar training patients are selected via cosine similarity. Their skip features are aggregated using normalized similarity weights:
\begin{equation}
\hat{s}^{(k)}_j = \sum_{i \in \mathcal{N}_K(j)} w_i \, s_i^{(k)}, \quad \text{where} \quad w_i = \frac{\max(\cos(u_j, z_i), 0)}{\sum_{m \in \mathcal{N}_K(j)} \max(\cos(u_j, z_m), 0)}.
\end{equation}
For efficiency, we use $K=1$ to supply spatial priors.
% For computational efficiency, we default to $K=1$, utilizing the single most analogous clinical neighbor to supply spatial priors.

\textbf{(III) Joint Objective Optimization.}
% During inference, the testing patient's synthetic MRI is generated by passing the adapted embedding and aggregated skip features into the frozen decoder: $\widehat{X}_j = f_{\mathrm{dec}}(\tilde{z}_j, \hat{s}^{(1)}_j, \dots, \hat{s}^{(4)}_j)$. 
During inference, the synthetic MRI is generated as: $\widehat{X}_j = f_{\mathrm{dec}}(\tilde{z}_j, \hat{s}^{(1)}_j, \dots, \hat{s}^{(4)}_j)$.
% To ensure the generated volumes are both anatomically realistic and diagnostically informative, the adapter $A_\theta$ and a lightweight classification head $C_\phi$ are jointly trained on the training cohort using a combined objective:
To ensure anatomical realism and diagnostic utility, the adapter $A\theta$ and classifier $C\phi$ are jointly trained with:
\begin{equation}
\mathcal{L}_{\text{total}} = \underbrace{\mathcal{L}_{\text{recon}}(X_i, \hat{X}_i)}_{\text{Reconstruction}} + \lambda_{\text{cls}} \underbrace{\mathcal{L}_{\text{cls}}(y_i, \hat{y}_i)}_{\text{Classification}}.
\end{equation}
% The classification term is a standard cross-entropy loss over the diagnostic labels, $\mathcal{L}_{\text{cls}} = \mathrm{CE}(y_j, C_\phi(\tilde{z}_j))$. The reconstruction loss combines $L_1$ distance, SSIM, and Total Variation (TV) regularization to enforce spatial smoothness and suppress artifacts:
The classification loss is cross-entropy, $\mathcal{L}{\text{cls}} = \mathrm{CE}(y_j, C_\phi(\tilde{z}j))$. Reconstruction combines $L_1$, SSIM, and Total Variation (TV) \cite{rudin1992nonlinear}:
\begin{equation}
\mathcal{L}_{\text{recon}} = \|X_j - \hat{X}_j\|_1 + (1 - \text{SSIM}(X_j, \hat{X}_j)) + \lambda_{\text{TV}} \cdot \text{TV}(\hat{X}_j).
\end{equation}
% By co-optimizing these objectives, the MIRAGE framework effectively synthesizes structural MRIs that preserve critical diagnostic biomarkers for patients lacking empirical neuroimaging data.
Joint optimization enables synthesis of structural MRIs preserving diagnostic biomarkers for patients without neuroimaging.

\section{Experiments}

\paragraph{\textbf{Datasets and Preprocessing.}}
% We utilized data from 1,175 patients from the ADNI dataset \cite{jack2008alzheimer}. Each patient record consists of 88-dimensional tabular EHR features, a structural brain MRI scan with dimensions of $91 \times 109 \times 91$, and a binary diagnostic label (Cognitively Normal [CN] or Alzheimer's Disease [AD]). The cohort was randomly split into training and testing subsets at a 7:3 ratio. To simulate real-world, EHR-only clinical scenarios, the true MRI scans of the testing patients ($\mathcal{P}_{\mathrm{te}}$) were masked during training. 
% For the knowledge graph, we employ the iBKH \cite{su2023biomedical}, comprising over 2 million entities and 48 million triples. SapBERT \cite{liu2021self} is used to generate semantic embeddings for patient-to-KG linking.
We used data from 1,175 ADNI patients~\cite{jack2008alzheimer}. Each record includes 88-dimensional tabular EHR features, a structural brain MRI of $91 \times 109 \times 91$, and a binary label (Cognitively Normal [CN] or Alzheimer's Disease [AD]). The cohort was randomly split into training and testing sets at a 7:3 ratio. To simulate real-world EHR-only scenarios, true MRIs of testing patients ($\mathcal{P}_{\mathrm{te}}$) were masked during training.
For the knowledge graph, we use iBKH \cite{su2023biomedical} with over 2 million entities and 48 million triples. SapBERT \cite{liu2021self} generates embeddings for patient-to-KG linking.

\paragraph{\textbf{Baselines.}}
% We compare \modelname\ against the following baselines: (1) EHR-only machine learning models (Logistic Regression, Random Forest, Multi-Layer Perceptron); (2) EHR$\to$MRI (direct conditional generation); (3) EHR$\to$Latent (MLP mapping to AE space); (4) MVAE (multimodal VAE with shared latents); and (5) Conditional Diffusion (EHR-conditioned iterative denoising).
We compare \modelname\ with: (1) EHR-only models (Logistic Regression, Random Forest, Multi-Layer Perceptron); (2) EHR$\to$MRI conditional generation; (3) EHR$\to$Latent MLP mapping to the AE space; (4) MVAE with shared latents; and (5) Conditional Diffusion with EHR-conditioned denoising.

\paragraph{\textbf{Implementation Details.}} 
% \textbf{MRI Processing:} Volumes were skull-stripped via FSL-BET, registered to MNI152 space ($1\,\mathrm{mm}$ isotropic), and intensity-normalized (z-score), resulting in $182 \times 218 \times 182$ voxels.
% \textbf{Downstream Classification:} To evaluate clinical utility, a 3D CNN classifier \cite{li2025transformer} is trained on real MRIs and evaluated on synthetic MRIs. We report Balanced Accuracy (BAcc), AUC, Sensitivity (SEN), and Specificity (SPE).
% \textbf{Fusion Protocol:} To assess modality complementarity, we train a logistic regression "fusion" module. It takes a 3D feature vector $[p_{\mathrm{EHR}}, p_{\mathrm{MRI}}, |p_{\mathrm{EHR}} - p_{\mathrm{MRI}}|]$ as input, where $p$ represents class probabilities from EHR-baselines and the synthetic-MRI-based 3D CNN, respectively.
% \textbf{MRI Processing:} Volumes were skull-stripped via FSL-BET \cite{jenkinson2012fsl}, registered to MNI152 space ($1,\mathrm{mm}$ isotropic), and z-score normalized, yielding $182 \times 218 \times 182$ voxels.
% \textbf{Downstream Classification:} A 3D CNN \cite{li2025transformer} is trained on real MRIs and evaluated on synthetic MRIs. We report Balanced Accuracy (BAcc), AUC, Sensitivity (SEN), and Specificity (SPE).
% \textbf{Fusion Protocol:} To assess modality complementarity, we train a logistic regression fusion module. It takes a 3D vector $[p_{\mathrm{EHR}}, p_{\mathrm{MRI}}, |p_{\mathrm{EHR}} - p_{\mathrm{MRI}}|]$ as input, where $p$ denotes probabilities from EHR baselines and the synthetic-MRI-based 3D CNN.
For MRI, volumes were skull-stripped via FSL-BET \cite{jenkinson2012fsl}, registered to MNI152 space ($1\mathrm{mm}$ isotropic), and z-score normalized, yielding $182 \times 218 \times 182$ voxels.
For downstream classification, a 3D CNN \cite{li2025transformer} is trained on real MRIs and evaluated on synthetic MRIs. We report Balanced Accuracy (BAcc), AUC, Sensitivity (SEN), and Specificity (SPE).
To assess modality complementarity, we train a logistic regression fusion module. It takes a 3D vector $[p_{\mathrm{EHR}}, p_{\mathrm{MRI}}, |p_{\mathrm{EHR}} - p_{\mathrm{MRI}}|]$ as input, where $p$ denotes probabilities from EHR baselines and the synthetic-MRI-based 3D CNN.

\paragraph{\textbf{Main Results.}}
Table~\ref{tab:main} presents performance comparisons across all baselines. Unimodal EHR-only classifiers show limited prognostic capacity, indicating that tabular clinical records obscure critical neurodegenerative trajectories when processed in isolation.
Standard generative baselines (Direct EHR$\to$MRI, MVAE, and Conditional Diffusion) fail to produce clinically viable outputs, showing mode collapse, structural corruption, or extreme blurring. These failures reflect an information-theoretic limitation: synthesizing complex, high-frequency 3D anatomical structures de novo from low-dimensional tabular data without spatial priors violates the Data Processing Inequality.
In contrast, \modelname\ circumvents this limitation by using the $K=1$ nearest-neighbor skip features as an anatomical canvas, while the aligned latent embedding ($\tilde{z}_j$) modulates disease-specific morphological changes. This biologically grounded synthesis generates clear, anatomically coherent MRIs that preserve individualized neurodegenerative biomarkers. When evaluated with a downstream 3D CNN, the reconstructed MRIs consistently amplify the prognostic value of the original EHR data. The \modelname-augmented fusion protocol improves the classification rate by 13\% over the strongest EHR-only baseline, showing that projecting clinical phenotypes into a structural manifold unlocks hidden diagnostic signals.

% \begin{table}[htbp]
\begin{table}[!t]
\captionsetup{font=scriptsize} 
\centering
\footnotesize
\caption{Performance comparison across methods. Bold and underlined values indicate the best and second-best results, respectively. Upper-bound rows use real MRI and EHR+MRI inputs for reference only and are not directly comparable. Our method achieves the best balanced accuracy and sensitivity when combined with an MLP classifier, while remaining competitive across AUC and specificity.}
\label{tab:main}
\scalebox{0.8}{%
\begin{tabular}{ccccc}
\hline
\textbf{Method} & \textbf{BAcc $\uparrow$} & \textbf{AUC $\uparrow$} & \textbf{SEN $\uparrow$} & \textbf{SPE $\uparrow$} \\ \hline
MRI Upper Bound     & 0.81 & 0.85 & 0.84 & 0.77 \\
EHR+MRI Upper Bound & 0.73 & 0.62 & 0.71 & 0.75 \\ \hline
Logistic Regression (LR)     & 0.66 & 0.75 & 0.58 & 0.75 \\
Multi-Layer Perceptron (MLP) & 0.62 & 0.72 & 0.42 & 0.82 \\
Random Forest (RF)           & 0.67 & \textbf{0.80} & 0.39 & \textbf{0.86} \\ \hline
EHR $\rightarrow$ MRI  & 0.65 & 0.72 & 0.52 & 0.80 \\
EHR-latent-decoder     & 0.64 & 0.72 & 0.48 & 0.81 \\
MVAE                   & 0.67 & 0.71 & 0.50 & 0.84 \\
Conditional Diffusion  & 0.66 & 0.67 & 0.47 & \underline{0.84} \\ \hline
Ours (LR+Syn MRI)  & 0.67 & 0.75 & \underline{0.56} & 0.78 \\
Ours (MLP+Syn MRI) & \textbf{0.70} & 0.75 & \textbf{0.63} & 0.78 \\
Ours (RF+Syn MRI)  & \underline{0.68} & \underline{0.78} & 0.55 & 0.81 \\ \hline
\end{tabular}
}
\end{table}

\begin{figure*}[!t]
\centering
\footnotesize
\captionsetup{font=scriptsize}

% ================= Table 1 =================
\begin{minipage}[t]{0.40\textwidth}
\centering
\captionof{table}{Ablation study results.}
\label{tab:ablation}
\renewcommand{\arraystretch}{1.2}
\scalebox{0.85}{
\begin{tabularx}{\linewidth}{lcccc}
\toprule
 & BAcc & AUC & SEN & SPE \\ \midrule
Ours & 0.70 & 0.75 & 0.63 & 0.78 \\
w/o KG & 0.54 & 0.56 & 0.27 & 0.81 \\
w/o GAT & 0.58 & 0.58 & 0.44 & 0.72 \\
w/o AE & 0.61 & 0.64 & 0.47 & 0.76 \\
w/o Adapter & 0.58 & 0.54 & 0.42 & 0.73 \\
\bottomrule
\end{tabularx}
}
\end{minipage}
\hfill
% ================= Table 2 =================
\begin{minipage}[t]{0.30\textwidth}
\centering
\captionof{table}{Human evaluation results.}
\label{tab:human_eval}
\renewcommand{\arraystretch}{1.6}
\scalebox{0.85}{
\begin{tabular}{lcc}
\toprule
Metric & Real & Synth. \\ \midrule
Anat. Real. & 0.54 & 0.42 \\
Clin. Usab. & 3.04 & 2.96 \\
Diag. Interp. & 3.00 & 2.98 \\
\bottomrule
\end{tabular}
}
\end{minipage}
\hfill
% ================= Figure =================
\begin{minipage}[t]{0.28\textwidth}
\centering
\captionsetup{type=figure}
\caption*{} % 占位对齐（不改间距，不缩小空白）
\includegraphics[width=\linewidth]{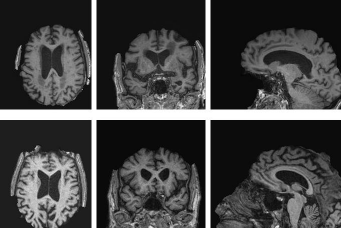}
\captionof{figure}{Case studies.}
\label{fig:case_study}
\end{minipage}

\end{figure*}

\paragraph{\textbf{Ablation Studies.}}
To isolate the contribution of each module and verify that the framework synthesizes patient-specific pathology rather than replicating cohort neighbors, we analyze four ablation settings (Table~\ref{tab:ablation}). The strong performance degradation in the \textit{w/o KG} (direct EHR-to-latent MLP) and \textit{w/o GAT} (nearest-neighbor feature transfer) settings validates that structured ontological mapping and dynamic message passing are essential for mitigating EHR sparsity and noise. Furthermore, the \textit{w/o AE} ablation, replacing the AE manifold with an MLP generator, causes performance collapse and confirms that a pre-trained population-level structural manifold is required to prevent generative hallucination. The decline in the \textit{w/o Adapter} setting (direct GAT-to-decoder decoding) highlights the need for the cross-modal alignment layer to bridge graph-structured clinical priors and latent image representations. The \textit{w/o Adapter} ablation also provides evidence against nearest-neighbor feature copying. In this setting, the decoder receives raw graph-propagated embeddings and $K=1$ spatial skip features. If the downstream 3D CNN relied on anatomical structures from clinical neighbors via skip connections, the performance drop would be marginal. Instead, diagnostic accuracy declines sharply when the adapter is removed. This confirms that high-frequency skip features provide non-diagnostic structural background, while disease-specific biomarkers such as ventricular expansion and hippocampal atrophy are synthesized from the aligned latent vector.
% ($\tilde{z}_j$).

% \begin{table}[htbp]

% \begin{table}[!t]
%     \centering
%     \footnotesize
%     \captionsetup{font=scriptsize} 
%     \begin{minipage}[t]{0.6\textwidth}
%         \centering
%         \caption{Ablation study results.}
%         \label{tab:ablation}
%         \renewcommand{\arraystretch}{1.2}
%         \scalebox{0.8}{
%         % \begin{tabularx}{\linewidth}{lcccc}
%         \begin{tabularx}{\linewidth}{lCCCC}
%             \toprule
%              & BAcc & AUC & SEN & SPE \\ \midrule
%             Ours & 0.70 & 0.75 & 0.63 & 0.78 \\
%             w/o KG & 0.54 & 0.56 & 0.27 & 0.81 \\
%             w/o GAT & 0.58 & 0.58 & 0.44 & 0.72 \\
%             w/o AE & 0.61 & 0.64 & 0.47 & 0.76 \\
%             w/o Adapter & 0.58 & 0.54 & 0.42 & 0.73 \\ \bottomrule
%         \end{tabularx}
%         }
%     \end{minipage}
%     \hfill
%     \begin{minipage}[t]{0.36\textwidth}
%         \centering
%         \footnotesize
%         \captionsetup{font=scriptsize} 
%         \caption{Human evaluation results.}
%         \label{tab:human_eval}
%         \renewcommand{\arraystretch}{1.75} 
%         \scalebox{0.8}{
%         \begin{tabularx}{\linewidth}{lcc}
%             \toprule
%             \textbf{Metric} & \textbf{Real} & \textbf{Synth.} \\ \midrule
%             Anat. Realism & 0.54 & 0.42 \\
%             Clin. Usability & 3.04 & 2.96 \\
%             Diag. Consistency & 3.00 & 2.98 \\ \bottomrule
%         \end{tabularx}
%         }
%     \end{minipage}
% \end{table}

\paragraph{\textbf{Case Study and Human Evaluation.}}

% To evaluate the clinical fidelity of the synthesized MRIs, we conducted a human assessment with two biomedical experts. We randomly sampled 50 scans (25 real, 25 synthetic) and evaluated them across three dimensions: 
% (i) Anatomical Realism (binary ``Yes/No'' rate for realism); 
% (ii) Clinical Usability (5-point scale for visual quality); and 
% (iii) Diagnostic Interpretability (5-point scale for structural clarity). 
% As shown in Table~\ref{tab:human_eval}, \modelname\ achieves high parity with real MRIs, with synthetic scans being nearly indistinguishable from ground truth in anatomical detail. Critically, the specialists confirmed that the synthetic MRIs consistently exhibit identifiable AD-related biomarkers, such as hippocampal atrophy and ventricular enlargement, that align with the patients' underlying clinical profiles. This expert validation confirms that our framework does not merely generate realistic images but successfully translates EHR-based disease patterns into clinically valid structural evidence, demonstrating its utility for downstream diagnostic support.

To evaluate the clinical fidelity of synthesized MRIs, we conducted a human assessment with two biomedical experts. We randomly sampled 50 scans (25 real, 25 synthetic) and evaluated three dimensions:
(i) Anatomical Realism (binary Yes/No rate);
(ii) Clinical Usability (5-point visual quality scale); and
(iii) Diagnostic Interpretability (5-point structural clarity scale).
As shown in Table~\ref{tab:human_eval}, \modelname\ achieves high parity with real MRIs, with synthetic scans nearly indistinguishable from ground truth anatomically. The specialists confirmed that synthetic MRIs consistently exhibit AD-related biomarkers, including hippocampal atrophy and ventricular enlargement, aligned with clinical profiles. This validation shows that the framework not only generates realistic images but also translates EHR-based disease patterns into clinically valid structural evidence for downstream diagnostic support.

% \begin{figure}[t]
% \centering
% \includegraphics[width=0.3\columnwidth]{resources/merged2.pdf}
% \caption{Case studies.}
% \label{fig:case_study}
% \end{figure}

% \begin{figure*}[!t]
% \centering
% \footnotesize
% \captionsetup{font=scriptsize}

% % ================= Table 1 =================
% \begin{minipage}[t]{0.38\textwidth}
% \centering
% \captionof{table}{Ablation study results.}
% \label{tab:ablation}
% \renewcommand{\arraystretch}{1.2}

% \begin{tabularx}{\linewidth}{lCCCC}
% \toprule
%  & BAcc & AUC & SEN & SPE \\ \midrule
% Ours & 0.70 & 0.75 & 0.63 & 0.78 \\
% w/o KG & 0.54 & 0.56 & 0.27 & 0.81 \\
% w/o GAT & 0.58 & 0.58 & 0.44 & 0.72 \\
% w/o AE & 0.61 & 0.64 & 0.47 & 0.76 \\
% w/o Adapter & 0.58 & 0.54 & 0.42 & 0.73 \\
% \bottomrule
% \end{tabularx}
% \end{minipage}
% \hfill
% % ================= Table 2 =================
% \begin{minipage}[t]{0.32\textwidth}
% \centering
% \captionof{table}{Human evaluation results.}
% \label{tab:human_eval}
% \renewcommand{\arraystretch}{1.6}

% \begin{tabular}{lcc}
% \toprule
% Metric & Real & Synth. \\ \midrule
% Anat. Real. & 0.54 & 0.42 \\
% Clin. Usab. & 3.04 & 2.96 \\
% Diag. Interp. & 3.00 & 2.98 \\
% \bottomrule
% \end{tabular}
% \end{minipage}
% \hfill
% % ================= Figure =================
% \begin{minipage}[t]{0.28\textwidth}
% \centering
% \captionof{figure}{Case studies.}
% \label{fig:case_study}
% \includegraphics[width=\linewidth]{resources/merged2.pdf}
% \end{minipage}

% \end{figure*}

\section{Discussion}
\label{sec:discussion}

The experimental results show that \modelname\ bridges the modality gap in AD diagnosis by synthesizing anatomically and clinically consistent MRIs from heterogeneous EHR data. Rather than a traditional feature engineering tool, the framework uses MRI synthesis as a generative objective forcing the model to distill high-dimensional clinical narratives into biologically grounded structural evidence. This process, termed generative feature distillation, leverages reconstruction of realistic brain anatomy, such as hippocampal volume or cortical thickness, to constrain the latent space and prioritize diagnostically useful features. Central to this success is the Biomedical Knowledge Graph (KG) as a semantic anchor, bypassing inconsistent EHR schemas by linking patients through standardized clinical entities. This ensures latent factors of neurodegeneration remain stable across cohorts, enabling transfer of imaging priors to EHR-only patients. By translating unstructured clinical observations into radiological evidence, \modelname\ improves predictive performance and interpretability via pathologically relevant biomarkers. Nevertheless, the study is limited by a lack of evaluation on external cohorts beyond ADNI. Although results show strong internal validity, future work will validate the framework on independent real-world EHR-only cohorts to assess generalization and robustness across clinical settings.
\section{Conclusion}

In this work, we introduced \modelname, a knowledge graph--guided cross-modal MRI synthesis framework designed to generate anatomically realistic and clinically meaningful MRIs for patients with only EHR data. By integrating 3D MRI autoencoding, knowledge graph--based embedding propagation, and adapter--decoder alignment, MIRAGE enables structure-consistent MRI generation without paired EHR–MRI training data. Experiments demonstrate that the synthesized MRIs preserve disease-relevant anatomy and enhance AD classification. Future work will extend \modelname\ to broader datasets and modalities to further assess its generalizability across clinical cohorts.

\bibliographystyle{splncs04}
\bibliography{main}
%





% LNCS Homepage, \url{http://www.springer.com/lncs}, last accessed 2023/10/25
% \end{thebibliography}
\end{document}